\documentclass[10pt,twocolumn,letterpaper]{article}

\usepackage{cvpr}
\usepackage{times}
\usepackage{epsfig}
\usepackage{graphicx}
\usepackage{amsmath}
\usepackage{amssymb}

\usepackage{bm}
\usepackage{subfigure}
\usepackage{multirow}
\usepackage{array}
\usepackage{authblk}
\usepackage{makecell}

\usepackage[bookmarks=false]{hyperref}
\cvprfinalcopy 

\def\cvprPaperID{842} 

\begin{document}

\title{FTGAN: A Fully-trained Generative Adversarial Networks for Text to Face Generation }

\author{Xiang Chen} 
\author{\;Lingbo Qing} 
\author{\;Xiaohai He} 
\author{\\Xiaodong Luo}
\author{\;Yining Xu}

\affil[1]{Sichuan University}
\affil[ ]{\tt\small {2016222050086@stu.scu.edu.cn}}
\renewcommand\Authands{, } 

\maketitle

\begin{abstract}
As a sub-domain of text-to-image synthesis, text-to-face generation has huge potentials in public safety domain.
With lack of dataset, there are almost no related research focusing on text-to-face synthesis.
In this paper, we propose a fully-trained Generative Adversarial Network (FTGAN) that trains the text encoder and image decoder at the same time for fine-grained text-to-face generation.  
With a novel fully-trained generative network, FTGAN can synthesize higher-quality images and urge the outputs of the FTGAN are more relevant to the input sentences. 
In addition, we build a dataset called SCU-Text2face for text-to-face synthesis. 
Through extensive experiments, the FTGAN shows its superiority in boosting both generated images' quality and similarity to the input descriptions. 
The proposed FTGAN outperforms the previous state of the art, boosting the best reported Inception Score to 4.63 on the CUB dataset.
On SCU-text2face, the face images generated by our proposed FTGAN just based on the input descriptions is of average 59\% similarity to the ground-truth, which set a baseline for text-to-face synthesis.

\end{abstract}


\vspace{-5pt}
\section{Introduction}
\vspace{-5pt}
Text-to-image synthesis is fundamental and novel research domain in computer vision, which was first proposed by Reed in 2016~\cite{reed2016generative}. 
It could be seen as a reverse task to image caption, aiming to generate natural images from input sentences. 
Similar to image caption, text-to-image synthesis helps to mining the relationship between text and image, exploring the visual semantic mechanism of human brain. Besides, it has huge application potentials in art creation, computer-aided design~\cite{xu2018attngan}, image searching and so on.

The classical methods for text-to-image synthesis mostly applied a similar framework. 
They utilize a pretrained text-encoder to encode the input descriptions as a semantic vector, then train a conditional GAN as image-decoder to generate natural images based on a vector combining the semantic vector and a noise vector which conforms to Normal Distribution.
Although such a framework could synthesize high-quality natural images, it split the training process of text-encoder and image-decoder.
In such a framework, the quality of the semantic vector encoded by text-encoder will dominate the best quality of the image-decoder process.
To tackle this issue, we build a fully-trained GAN(FTGAN) for text-to-image synthesis, which could train the text-encoder and image-decoder at the same time.

\begin{figure}[tb]
	\small
	\centering
	\begin{center}
		\includegraphics[width=1.0\linewidth]{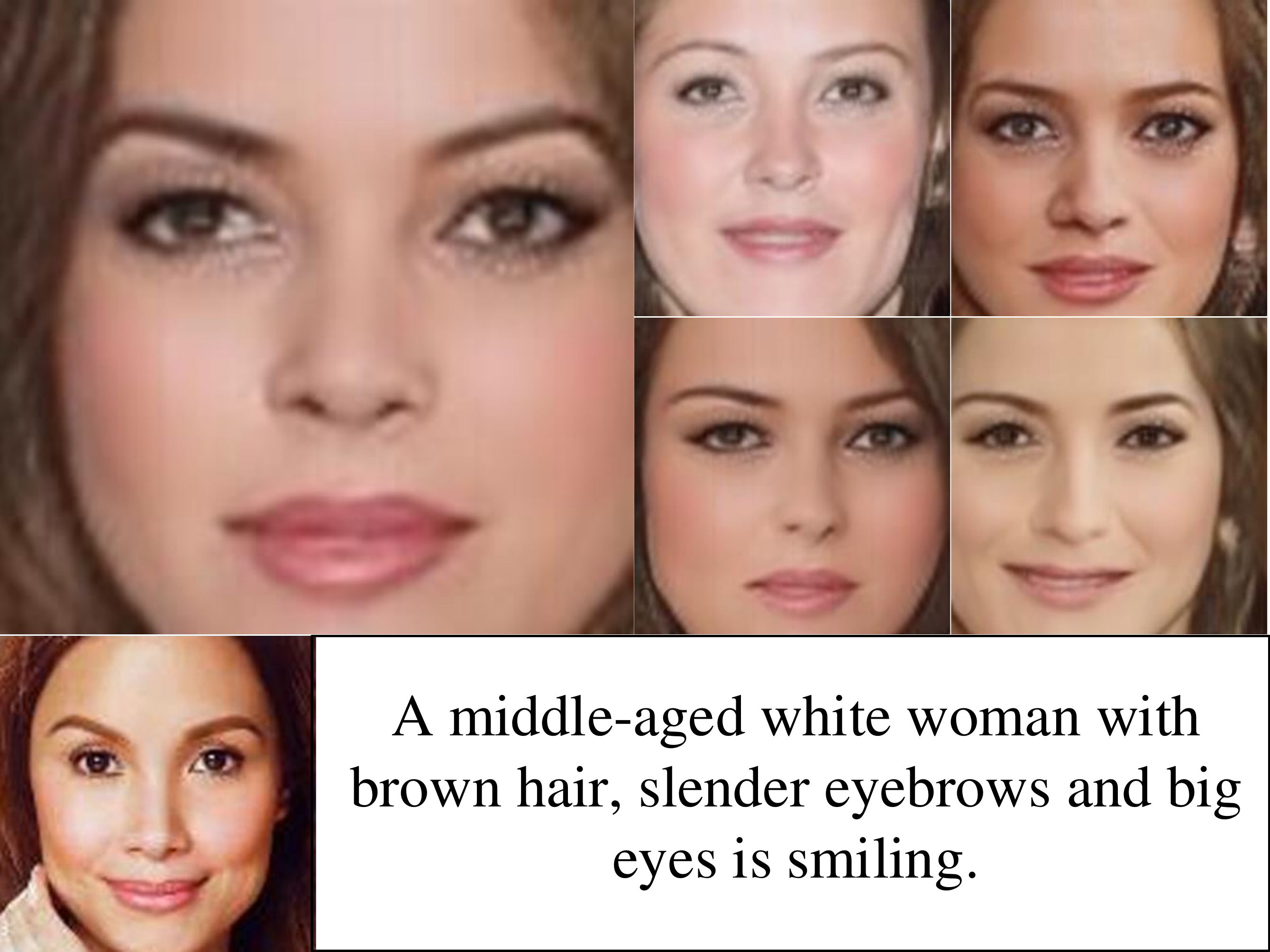}
	\end{center}
	\vspace{-10pt}
	\caption{An example of text-to-face synthesis. The 5 face images above are generated from the same sentence which is shown in the right down. And the face image in the left down is the ground-truth for the input sentence.
	}
	\vspace{-10pt}
	\label{fig:face_show}
\end{figure}

Text-to-face synthesis is a sub-domain of text-to-image synthesis, aiming to synthesize face images based on human descriptions. 
Similar to text-to-image synthesis, there are two main targets for text-to-face synthesis: (1) to generate high-quality images;(2) to generate images which are conformed to the input descriptions. This task, compared with text-to-image synthesis, has more relative values in the public safety domain. 
As we all know, drawing a picture for suspect just based on the descriptions of the eyewitnesses is a difficult task, which requires professional skills and rich experience. And it is also time-consuming. 
However, with a well-trained text-to-face model, a normal person could directly generate photo-realistic faces of suspects based on the descriptions of eyewitnesses quickly. 

For text-to-image synthesis, the common datasets is CUB~\cite{wah2011the}, Oxford102~\cite{nilsback2008automated} and COCO~\cite{lin2014microsoft}.
Since text-to-face is a sub-domain of text-to-image synthesis, those state-of-the-art networks can be also applied in text-to-face synthesis. 
However, there are few research are focused on text-to-face synthesis with no standard text-to-face datasets available. 
This is because there are no standard datasets for text-to-face synthesis. 
To our best know, there are some research focus on text to face sketch synthesis~\cite{di2018face} and attributes vector to sketch to natural face synthesis~\cite{Text2Sketch}. However, for generating natural faces from descriptions, there is only a repository named T2F on Github(https://github.com/akanimax/T2F), which build a network based on ProGAN~\cite{karras2018progressive} and StackGAN~\cite{zhang2017stackgan} and utilized a dataset called Face2text~\cite{gatt2018face2text} for training and testing. 
But its synthesized results are of poor quality.
To tackle this issue, we build a dataset SCU-Text2face based on CelebA~\cite{liu2015deep}, which contains 1000 images.
For each face images in SCU-Text2face, there are five descriptions given by different persons. This dataset could help to build a baseline for text-to-face synthesis task.

The main contribution of our method is threefold. 
(\textit{i}) An Fully-trained Generative Adversarial Network FTGAN is proposed for synthesizing images from text descriptions.  Experimental results show that the FTGAN significantly outperforms previous state-of-the-art GAN models. 
(\textit{ii}) A text-to-face dataset SCU-Text2face is build for text-to-face synthesis task. 
(\textit{iii}) A baseline for text-to-face synthesis is build based on FTGAN. To our best know, it is the first research focusing on generating natural faces from text descriptions.

\section{Related Work}
There are two main domains are related to text-to-face synthesis: (1)text-to-image synthesis; (2)face generation. Though there are few research focusing on text-to-face synthesis, it can benefit much from the development of this two domains.
\vspace{-5pt}
\subsection{Text-to-image Synthesis} \label{sec:text2image}
Despite there are kinds of networks for text-to-image synthesis, they are mostly based on encoder-decoder framework and conditional GAN~\cite{mirza2014conditional}. 
This encoder-decoder framework inludes text-encoder and image-decoder. 
The text-encoder turn input descriptions to semantic vectors and the image-decoder turn the encoded semantic vectors to natural images.
There are two main targets for text-to-image synthesis: to generate high-quality images and generate images matching the given descriptions. 
All the developments of text-to-images synthesis are based on this two targets.

The early research for text-to-image synthesis are mainly focusing on improving the quality of generated images.
The task text-to-image was first presented in 2016, Reed \etal presented this novel task and developed two end-to-end networks based on conditional GAN to accomplish it~\cite{reed2016generative}. Reed utilized a pretrained Char-CNN-RNN network for text encoding and built a network similar to DCGAN~\cite{radford2016unsupervised} as image decoder to generate natural images from vector.
Then many researchers made some progresses based on his work~\cite{dong2017semantic}. One of the most influential research is made by Zhang \etal, they proposed a 2-stages network StackGAN to solve this task, which could generate high-quality images and improved the Inception Score obviously~\cite{zhang2017stackgan}. This network is also inherited by later research~\cite{zhang2018stackgan++,xu2018attngan,zhang2018photographic,qiao2019mirrorgan}.

Since the network has already been capable to generate realistic images, researchers progressively focused on achieving another target: improving the similarity between input text and generated images. 
Reed \etal proposed a network to generate images based on a box which was first generated. This method helped to generate more accurate results on the output images~\cite{reed2016learning}. 
Hong \etal also designed a GAN network based on a similar idea~\cite{hong2018inferring}. 
On the other side, Sharma \etal utilized dialog to assist the understanding for the description, which helps to synthesize images more relative to the input text~\cite{sharma2018chatpainter}. 
Dong \etal proposed an approach to generate new images based on the input image and descriptions, which can generate new images which matching input descriptions~\cite{dong2017semantic}. 
Besides, they also proposed a new training method called Image-Text-Image (I2T2I) which integrates text-to-image and image-to-text (image captioning) synthesis to improve the performance of text-to-image synthesis~\cite{dong2017i2t2i}.
Attention mechanisms have already achieved great breakthroughs in text-related and image-related tasks~\cite{yang2016stacked,xu2015show,zhang2018self-attention,vaswani2017attention}, now it also being used in GANs for text-to-image generation, Xu \etal~\cite{xu2018attngan} built AttnGAN firstly develops an attention mechanism that enables GANs to generate fine-grained high resolution images from nature language description. 
Qiao \etal~\cite{qiao2019mirrorgan} proposed a text-to-image-to-text network called MirrorGAN which applied a global-local collaborative attention model.
Since there is no available criterion how the generated images matching to the input descriptions, Zhang \etal~\cite{zhang2018photographic} proposed a visual-semantic similarity measure as an assist to evaluation metrics.
Those research imply a trend that researchers are progressively focusing on boosting the consistency between generated images and input sentences.

%

\subsection{Face Synthesis} \label{sec:face synthesis}

Since GAN was proposed by Goodfellow in 2014~\cite{goodfellow2014generative}, image synthesis has been a hot topic in deep learning. 
Because there are two large scale public dataset: CelebA and LFW~\cite{LFWTech}, face synthesis is also a popular research domain.
Almost most of the state-of-art networks will examplify their model's superiority on face synthesis, including networks based on GAN and networks based both on conditional GAN(such as DCGAN~\cite{radford2016unsupervised}, CycleGAN~\cite{zhu2017unpaired}, ProGAN~\cite{karras2018progressive}, BigGAN~\cite{brock2018large}, StyleGAN~\cite{styleGAN}, Stargan~\cite{choi2018stargan} and so on). 
With the development of those networks, the quality of generated face images are becoming better and better. 
Now some networks could even generate 1024$\times$1024 face images, much larger than the original images resolution of the face dataset. 
Those models aim to learn a mapping from noise vector which conforms to Normal distribution to natural face images. But they can't control the network to generate a precise face image which they want.

To tackle this issue, with conditional GAN, face synthesis have derived many interesting applications about face, such as translating edges to natural face images~\cite{wang2018highresolution}, exchanging the attributes of two face images~\cite{bao2018towards}, generating a positive face from the side face~\cite{huang2017beyond}, generating a full face from eyes' region only~\cite{eye2face}, from face attributes to sketches to natural face images synthesis~\cite{di2018face},face inpainting~\cite{yu2018generative} and so on. 
Those networks try to control the synthesized face images by adding a condition vector, could generate face images which meet the needs of different situations.
Text-to-face synthesis is similar to those tasks, which utilize the input descriptions as the control condition.


\section{Fully-trained Generative Adversarial Network}

In this section, we will elaborate the framework and details about FTGAN. 
At first, we will compare the framework of FTGAN with the previous text-to-image network. 
Then a comprehensive description of the network design of FTGAN will be given.

\subsection{Fully-trained text-to-image framework} \label{sec:Fattn}
\vspace{-5pt}
The framework of text-to-image synthesis could be divided into two parts: text-encoder and image-decoder.
Text-encoder is responsible for encoding the input sentences to semantic vectors, the Char-CNN-RNN~\cite{reed2016learning} used in Reed's work could be seen as a text-encoder. 
Image-decoder is to generate natural images based on the semantic vectors encoded by text-encoder, which is often similar to networks like DCGAN.
Current GAN-based models for text-to-image generation~\cite{reed2016generative,reed2016learning,zhang2017stackgan,zhang2018stackgan++,xu2018attngan} typically split the training of text-encoder and image-decoder. 
They trained text-encoder firstly, and then utilize the pre-trained text-encoder to train the image-decoder.
Different from most of the previous networks, AttnGAN designed a DAMSM network to do text-encoding and calculate the attention map, instead using the Char-CNN-RNN for encode the input sentences. Our work is mainly based on this network.
In this section, we propose a novel framework for text-to-image synthesis, which train the text-encoder and image decoder at the same time.

\begin{figure}[tb]
	\small
	\centering
	\begin{center}
		\includegraphics[width=1.0\linewidth]{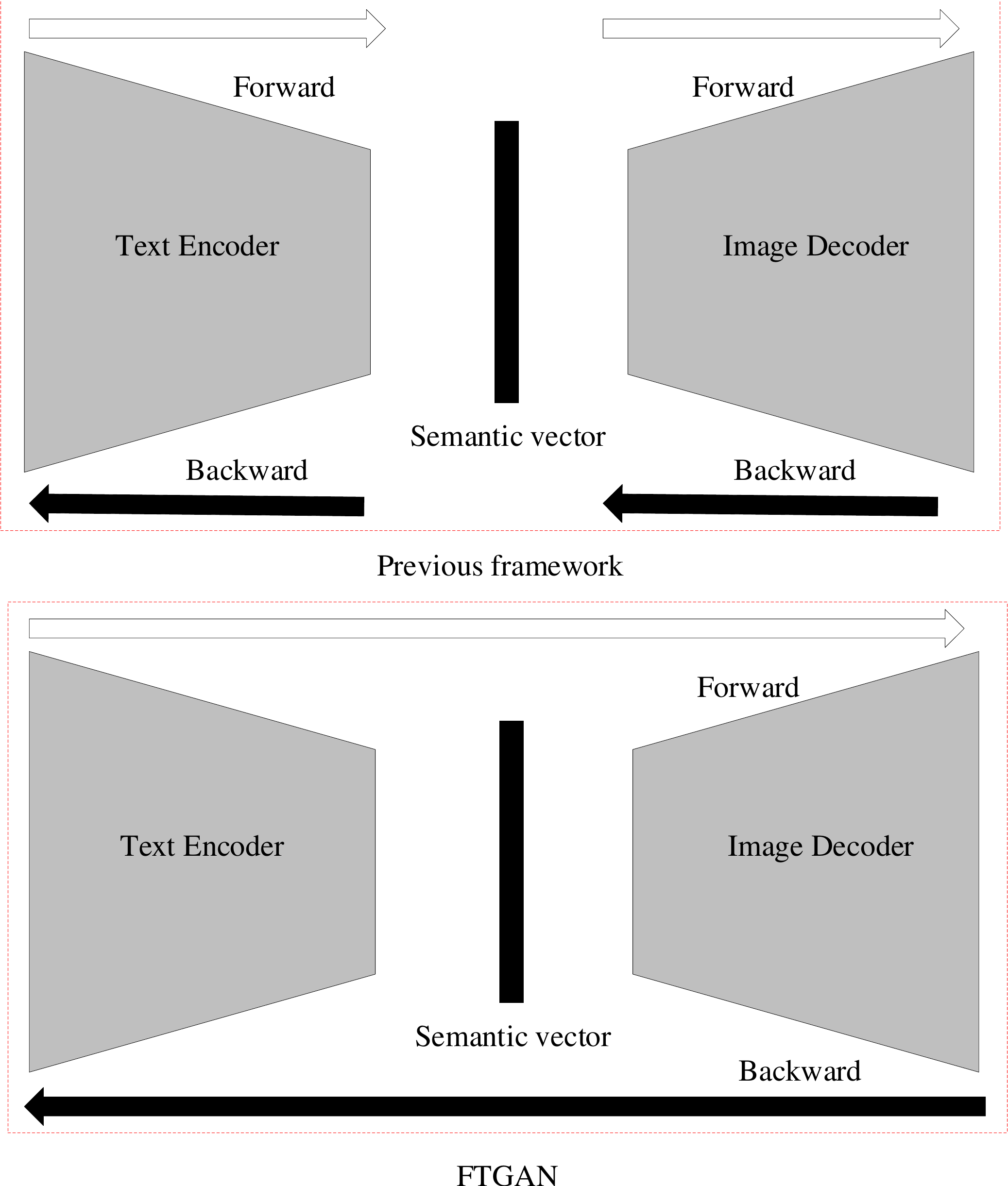}
	\end{center}
	\vspace{-10pt}
	\caption{The framework comparison between the proposed FTGAN and previous frameworks. Previous text-to-image synthesis networks train the text-encoder and image-decoder separately, text-encoder train a network from input sentences to semantic vectors and image-decoder train a network from semantic vectors to synthesized images.  However, FTGAN train a network directly from input sentences to the synthesized images.
	}
	\vspace{-10pt}
	\label{fig:framework}
\end{figure}

As is shown in Figure~\ref{fig:framework}, the common networks for text-to-image synthesis are based on an encoder-decoder framework. The encoder takes sentences as input and encode it to a semantic vector. The decoder then turn this semantic vector to a natural image. This two parts are of equal importance to text-to-image task. 
However, most previous research split this framework to two networks and train them separately. 
Reed firstly proposed a network to solve text-to-image task, they used a pre-trained network called Char-CNN-RNN to calculate the semantic vector of the input text, and then utilized a CNN similar to DCGAN to generate image with this semantic vector. 
When training this network, they actually just train the CNN network and split the training between encoder and decoder(previous framework in Figure~\ref{fig:framework}). Later research are mostly based on this framework and try to improve the efficiency of the CNN. 

However, as the base for image-decoder, the effect of the pre-trained text encoder will directly determined the upper limit of image-decoder. 
There are two main tasks for text-to-image synthesis: generating high-quality images; the images should conform to the meaning of input text. 
Using a pre-trained text-encoder and just training the image-decoder could generate high-quality images to some extent.
However, we couldn't make sure if generated images are what we want, because the input of the image-decoder is the semantic vector, which are highly determined by the pre-trained text-decoder. 
In order to generate higher-quality images which further matching the input text, we should train the text-encoder and image-decoder at the same time.

\vspace{-5pt}

\subsection{Fully-trained Generative Adversarial Networks Design} \label{sec:DAMSM}

\vspace{-5pt}
\begin{figure*}[tb]
	\begin{center}
		\includegraphics[width=1.0\linewidth]{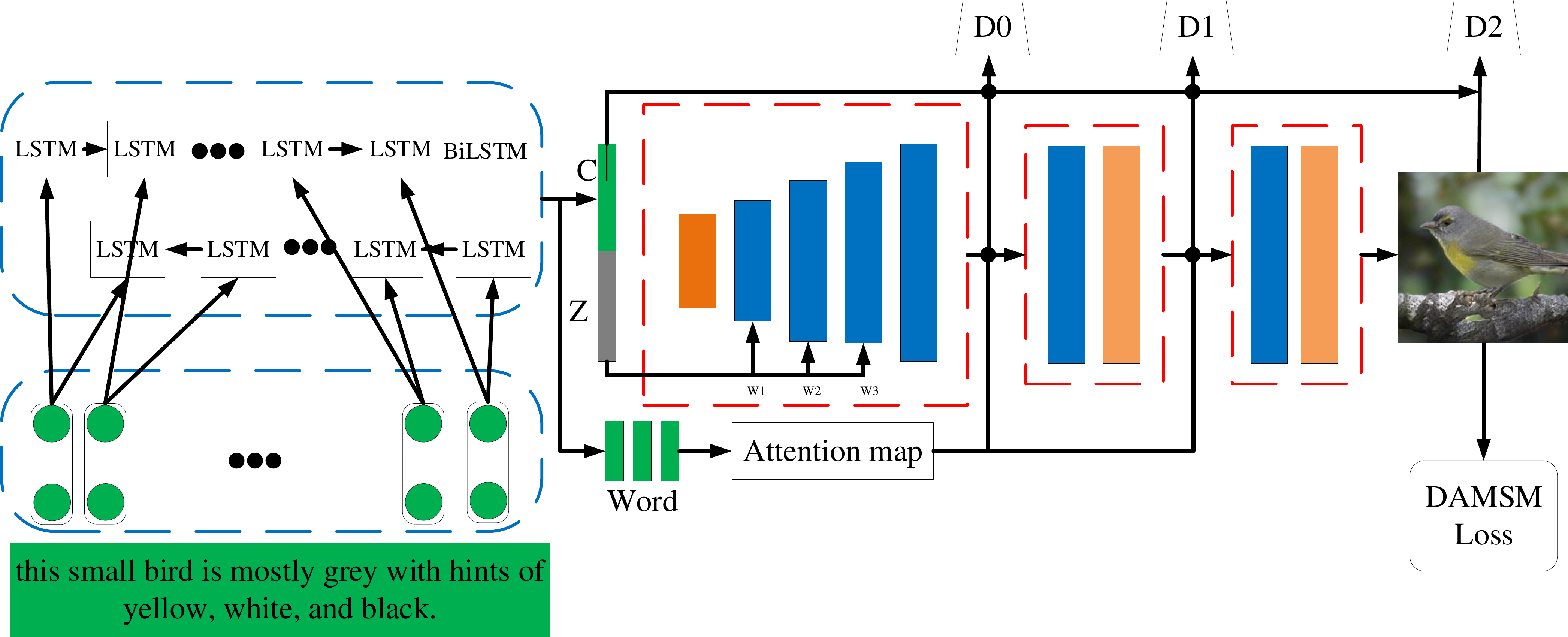}
	\end{center}
	\vspace{-10pt}
	\caption{The architecture of the proposed FTGAN. The left part in the blue box is the text-encoder and right part in the red box is the image-decoder. The DAMSM loss is same as the DAMSM loss in AttnGAN, which is part of the generator loss.
	}
	\vspace{-10pt}
	\label{fig:network}
\end{figure*}
%
%
In this section, we will describe the network details of the proposed FTGAN.  
Figure~\ref{fig:network} shows the detail network of the proposed fully-trained generative adversarial network (FTGAN). The main network are based on conditional GAN, include one generator and 3 discriminators (for different scale at 64$\times$64, 128$\times128$ and 256$\times$256). 
The generator is an encoder-decoder network, which is the main part of text-to-image synthesis.
It can be divided into two parts: text-encoder and image-decoder. 
In previous text-to-image networks, the image-decoder is the most important part. 
From StackGAN to AttnGAN and MirrorGAN, multi-stages image-decoder has proved its superiority in genarating high-quality images. 
Here, we also follow this idea.
In the follow sections, we will describe main parts of the proposed FTGAN separately.

\textbf{The text-encoder }{
is construted by a bi-directional Long Short-Term Memory (BiLSTM) that extracts semantic vectors from the input descriptions. 
In the BiLSTM, each word corresponds to two hidden states, one for each direction. 
We concatenate its two hidden states to represent the semantic meaning of each word. 
Through the text-encoder, the input sentences will be encoded as a matrix of $e \in \mathbb{R}^{D\times T}$. 
Its $i^{th}$ column $e_i$ is the feature vector for the $i^{th}$ word. 
$D$ is the dimension of the word vector and $T$ is the number of words.
On one side, the sentences embedding will be used for calculating the attention maps, which are the inputs of the last two stages in image-decoder, helping to guide the image generation process.
Meanwhile, the last hidden states of the BiLSTM are concatenated to be the global sentence vector, denoted by $C \in \mathbb{R}^{D}$. 
The semantic vector $C$ will be concated with a noise (conform to normal distribution) to a new vector, which is the input of image-decoder. 
To boost the stability of training process, we pre-trained the text-encoder network first.
However, different from previous networks, the parameters in text-encoder will be also updated when training the image-decoder.
}

\textbf{The image-decoder }{
is a 3-stages Convolutional Neural Network (CNN) that maps semantic vectors to natural images. 
The first stage takes the vector C generated by text-encoder concated with a noise (conforms to normal distribution) as input, reshaping it into 4$\times$4 feature maps(the dark yellow block in Figure~\ref{fig:network}). 
Through 4 upsample blocks(blue blocks), the 4$\times$4 feature maps will be enlarged to 64$\times$64.
The upsample block is a deconvolution layer, each will enlarge the scale of the feature map twice as it before.
Follow the noise processing in StyleGAN~\cite{styleGAN}, the input noise vector Z will not only be utilized to be combined with semantic vector C, it is also be weighted ($W1,W2,W3$ respectively) added into the first 3 deconvolution layers(at 8$\times$8, 16$\times$16, 32$\times$32 scale) after full connected layer and reshape operations.

The second and third stages are similar, which are both consist by an upsample block.
Different from the upsample blocks in the first stage, the upsample block in the next two stages are followed by a finetune block(light yellow blocks), which is used for further tuning the feature maps after upsampling. 
The fine-tune block is a constructed by a convolutional layer with a 3$\times$3 kernel.
The second parts take the 64$\times$64 feature maps and attention maps as input, and generate 128$\times$128 images. 
The third part is similar to the second part, the only difference is that the feature maps scale is from 128$\times$128 to 256$\times$256. 
The attention maps are calculated by referring to the attention maps in AttnGAN, and there is one feature map for every words of the input sentences.
After two upsample blocks in the next two stages, 256$\times$256 images will be generated, which will be used for calculating a generator loss. 
}

\textbf{The discriminators }{
in the FTGAN are similar to each other, referring to previous networks~\cite{reed2016learning,xu2018attngan}.
D0, D1 and D2 all takes sentence embedding C and its corresponding generated images(64$\times$64, 128$\times$128, 256$\times$256 respectively) as inputs.
The input images will firstly be downsampled to 4$\times$4 feature maps by several downsample blocks(according to the resolution of input images).
Each downsample blocks contains a convolution layer, a batchnormalization layer and a leaky relu layer.
Then the sentence embedding vector $C$ will also be reshaped to the same shape as image feature maps after being reshaped and repeated.
The image feature maps and sentence feature maps will be concated.
After several convolution layers, we get the final outputs of discriminator.
For the ground-truth and semantic vector pair, discriminator should define it as true. And for the generated images and semantic vector pair, discriminator should define it as false.
} 

\textbf{Loss function}{is also an important part of text-to-image synthesis.
The loss functions of FTGAN includes generator loss and discriminator loss.
The total generator loss is divided into two parts: the original generator loss and DAMSM loss. 
The generator loss $\mathcal{L}_{Gi}$ is similar to common CGAN's generator loss, include conditional part and unconditional part. 
But it calculate the generator loss at 3 scales (64, 128 and 256 respectively). And the DAMSM loss is calculated by a pre-trained DAMSM~\cite{xu2018attngan}.
The image-decoder generate images for every stage at different scales: 64$\times$64, 128$\times$128 and 256$\times$256. 
Every output images of the 3 stages will be used to calculate generator losses $\mathcal{L}_{stage_{i}}$:
\begin{equation}\label{eq:hybrid-LGi}
\footnotesize
\begin{aligned}
\mathcal{L}_{stage_{i}} &= -\frac{1}{2} \mathbb{E}_{\hat{x}_{i} \sim {p_{G_{i}}}} [\log(D_{i}(\hat{x}_{i})] -\frac{1}{2} \mathbb{E}_{\hat{x}_{i} \sim {p_{G_{i}}}} [\log(D_{i}(\hat{x}_{i}, C)],
\end{aligned}
\end{equation}
where $\hat{x}_{i}$ is the generated images for every stage,$D_{i}()$ is the $i^{th}$ discriminator,$C$ is the semantic vector of input sentences.

In order to boost the similarity between input sentences and output images, DAMSMS loss in AttnGAN is also used in generator to guide the training process. 
Therefore, the total generator loss is:
\begin{equation}\label{eq:final-LG}
\mathcal{L_G} = \mathcal{L}_{stage} + \lambda\mathcal{L}_{DAMSM}, \; \; \text{where} \; \mathcal{L}_{stage} = \sum_{i=0}^{m-1} \mathcal{L}_{stage_{i}},
\end{equation}

The discriminator losses are also similar to common discriminator loss, include conditional loss and unconditional loss.
Because AttnGAN has done a great job in the image-decoder of text-to-image network, in the part of image-decoder we mainly refer to this network. 
However, the kernel idea of FTGAN is to train the text-encoder and image-decoder at the same time, which could help to mine deeper relations between text and images, finally generating higher-quality and higher-semantic similarity images.
For every discriminator, the discriminator loss $\mathcal{L}_{D_{i}}$ is:

\begin{equation}\label{eq:hybrid-LDi}
\scriptsize
\begin{aligned}
\mathcal{L}_{D_{i}} &= -\frac{1}{2}  \mathbb{E}_{x_{i} \sim {p_{data_i}}} [\log D_{i}(x_{i})] \; -\frac{1}{2} \mathbb{E}_{\hat{x}_{i} \sim {p_{G_{i}}}} [\log(1 - D_{i}(\hat{x}_{i})]  + \\   
&\;\;\;-\frac{1}{2}  \mathbb{E}_{x_{i} \sim {p_{data_i}}} [\log D_{i}(x_{i}, C)] \; -\frac{1}{2} \mathbb{E}_{\hat{x}_{i} \sim {p_{G_{i}}}} [\log(1 - D_{i}(\hat{x}_{i}, C)] ,
\end{aligned}
\end{equation}
where $x$ is the ground-truth of the input description. The 3 discriminators are optimized independently.

In summary, we propose novel framework for text-to-image task, which train a total network from input sentences to output images, combining the text-encoder with image-decoder.
The fully-trained mechanism enables the network update the parameters in both text-encoder and image-decoder at the same time, which helps to boost the consistency between input sentences and generated images and improve the quality of final synthesized images.
}

\section{Experiments}
\vspace{-5pt}
{
In this section, extensive experiments are carried out to evaluate the proposed FTGAN. 
We first exemplified the superiority of our proposed FTGAN by comparing with the previous state-of-the-art GAN models for text-to-image~\cite{zhang2017stackgan,zhang2018stackgan++,reed2016generative,reed2016learning} on public dataset CUB~\cite{wah2011the}.
Then, we further prove the efficiency of FTGAN on SCU-Text2face, comparing it with AttnGAN and building a baseline for text-to-face synthesis task.

The poposed network is trained on a single 1080Ti GPU. In all our experiments, we empirically set $\lambda=5.0$ for ${L}_{DAMSM}$.
}

\subsection{SCU-Text2face Dataset Construction}  \label{sec:SCU-text2face}
\begin{figure}[tb]
	\begin{center}
		\includegraphics[width=1.0\linewidth]{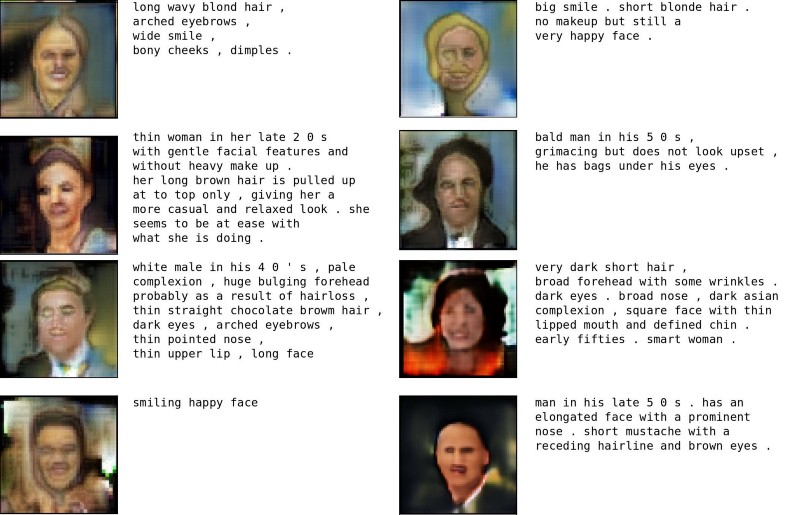}
	\end{center}
	\vspace{-10pt}
	\caption{The generated example of the T2F project on the Github. They proposed a network based on ProGAN and StackGAN, and used a dataset called Face2text~\cite{gatt2018face2text} for training and testing.
	}
	\vspace{-10pt}
	\label{fig:t2f}
\end{figure}

In public safety domain, the task of text-to-face is of huge potentials. However, because of lack of dataset, there are few research focus on this task. 
To our best know, there are just a project on Github and a conference paper are focus on this task. 
{\href{https://github.com/akanimax/T2F} {The project T2F}} on Github designed a network based on ProGAN and StackGAN, using the Face2text dataset for training and testing. 
But the results of this project are not so satisfactory (as shown in Figure~\ref{fig:t2f}). 
As for the conference paper, what we could only found is an abstract. 
Therefore, there are still no satisfactory baseline for text-to-face task.

\begin{figure}[tb]
	\begin{center}
		\includegraphics[width=1.0\linewidth]{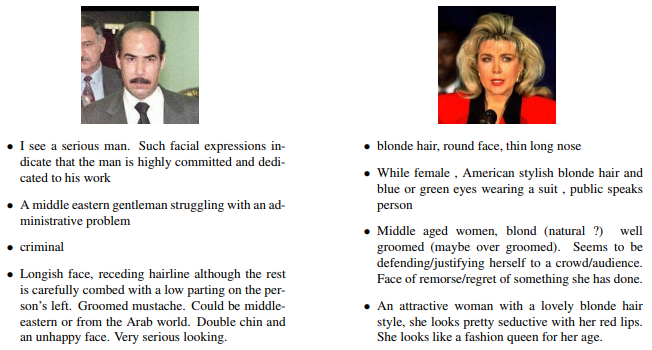}
	\end{center}
	\vspace{-10pt}
	\caption{Some examples of Face2text~\cite{gatt2018face2text}, which contains 400 samples. For each face images in Face2text, there are 5 related descriptions. But the descriptions are given freely, some of them even just contains one word(shown in the third description of the left face image).
	}
	\vspace{-10pt}
	\label{fig:Face2text_dataset}
\end{figure}

\begin{figure}[tb]
	\begin{center}
		\includegraphics[width=1.0\linewidth]{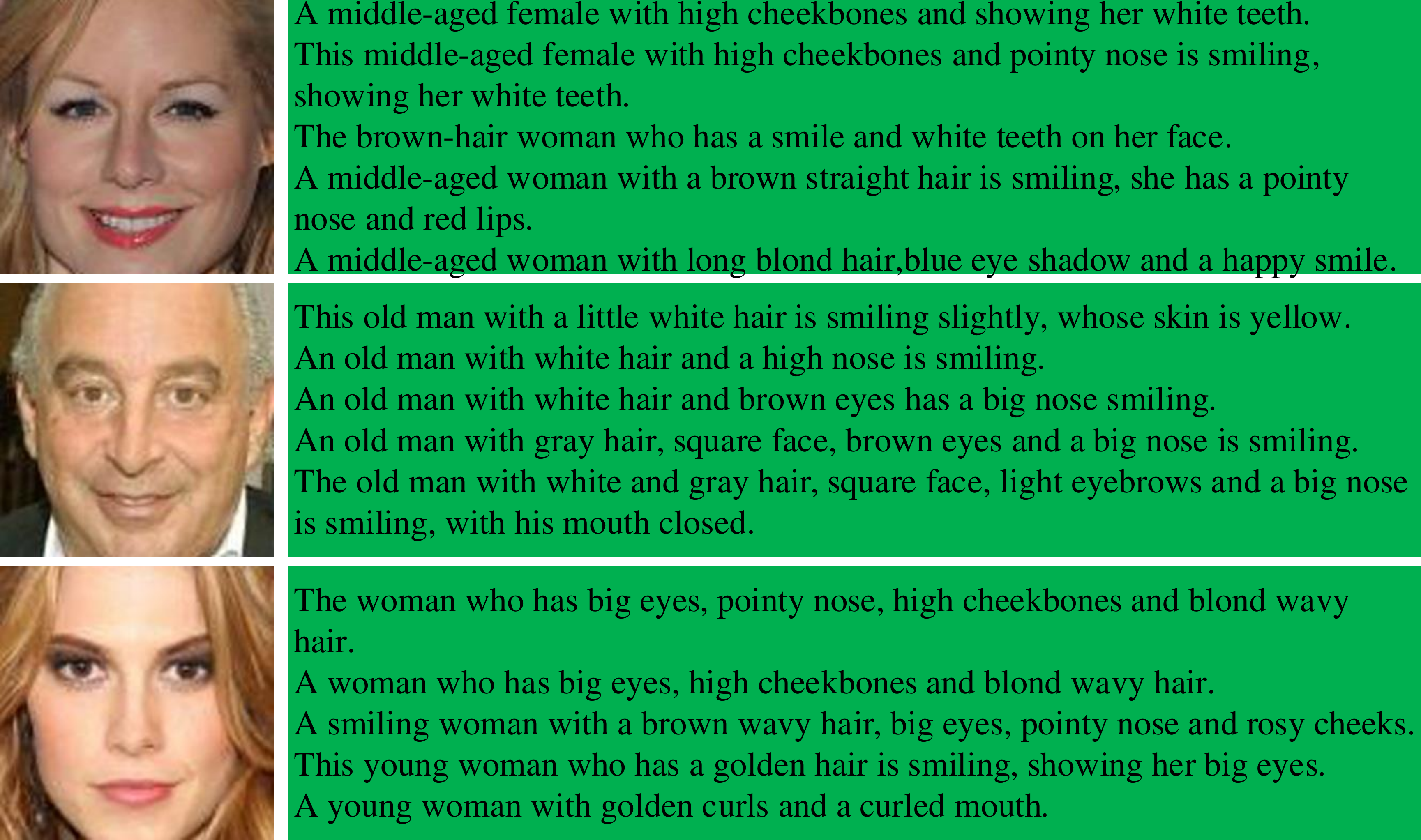}
	\end{center}
	\vspace{-10pt}
	\caption{Some examples of SCU-Text2face. Face images are selected from CelebA and normalized to 256$\times$256. There are 5 descriptions for every samples in SCU-Text2face.
	}
	\vspace{-10pt}
	\label{fig:SCU-Text2face}
\end{figure}

The Face2text dataset is a dataset originally used for image caption. 
Just like CUB and COCO, it could also be used for text-to-face synthesis. 
However, this dataset only contains 400 images and the descriptions for those images are not very formal (as shown in Figure~\ref{fig:Face2text_dataset}).
Referring to public dataset CUB and COCO, we build a dataset called SCU-Text2face for text-to-face synthesis based on the public face dataset CelebA. 
SCU-Text2face contains 1000 face images. 
For each of the face images in it, there are 5 descriptions from 5 different persons.
To build a standard text-to-face dataset, we firstly selected 1000 images from CelebA, which all belong to different persons. 
To maintain a balance of the dataset, those face images in SCU-Text2face contains people who have different ages, sexes and skins. 
For normalization, all the face images are cropped and reshaped into 256 $\times$256. 
Figure~\ref{fig:SCU-Text2face} shows some example of the SCU-Text2face.

\vspace{-5pt}

\subsection{Experimrnts on CUB} \label{sec:component}
\vspace{-5pt}

\begin{figure*}[tb]
	\begin{center}
		\includegraphics[width=1.0\linewidth]{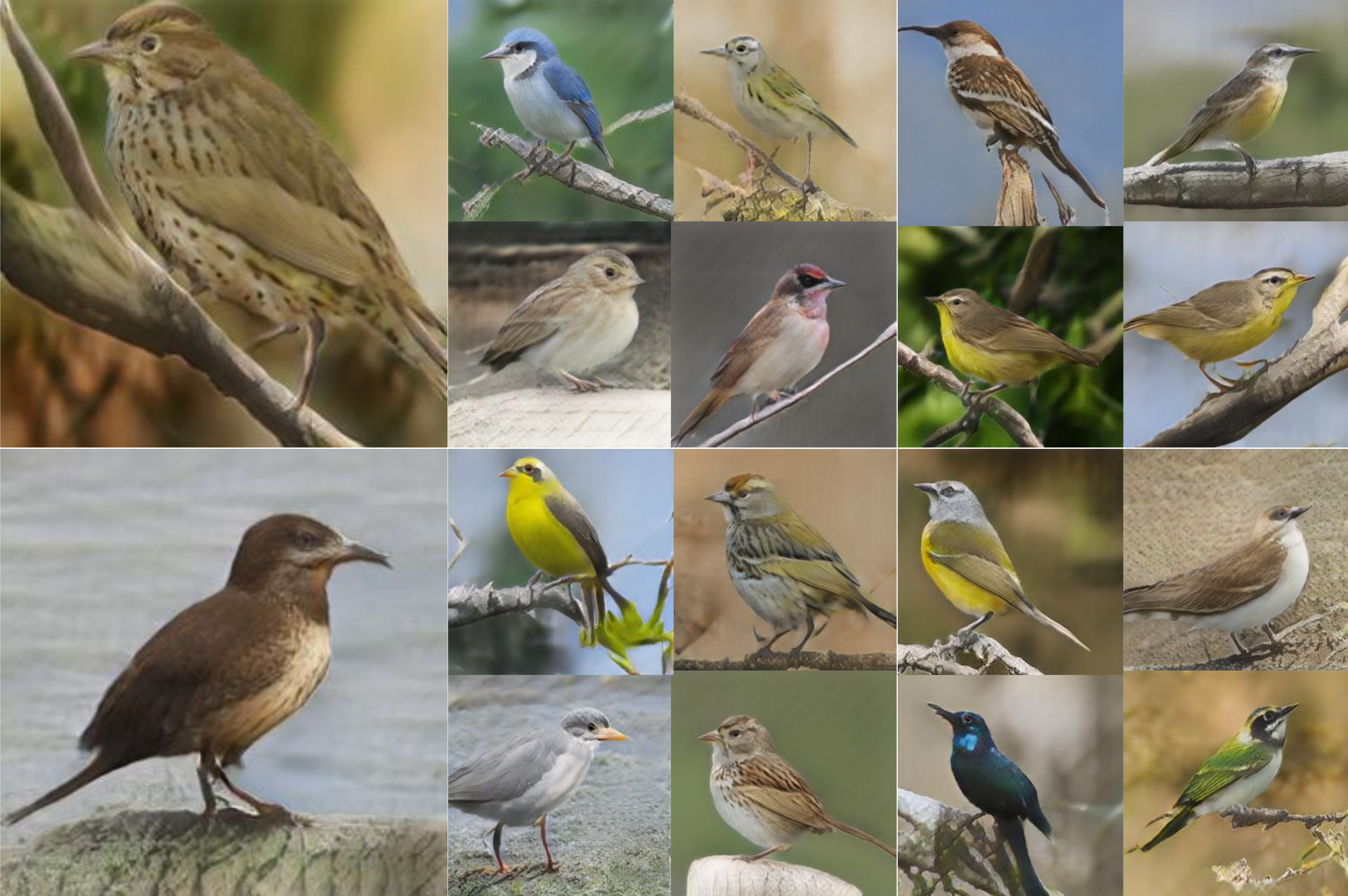}
	\end{center}
	\vspace{-10pt}
	\caption{Some examples of the generated images by FTGAN on CUB dataset.  
	}
	\vspace{-10pt}
	\label{fig:CUB_results}
\end{figure*}

To prove the superiority of FTGAN, we will first evaluate the proposed network on public dataset CUB. 
CUB is one of the most popular dataset in text-to-image synthesis(the other two are Oxford102 and COCO), includes 200 birds species. 
Oxford102 is a dataset of flowers, which is similar to CUB but contains fewer images.
The scale of COCO is much larger than CUB and Oxford102, experiments on which is very time-consuming.
Thus, we finally choose the CUB dataset to exemplify the proposed FTGAN.
Follow the preprocess in previous research~\cite{zhang2017stackgan,zhang2018stackgan++,reed2016generative,reed2016learning}, we divide this dataset into training set (180 species) and test set(20 species), contains 8855 and 2933 images respectively.
In this section, we first quantitatively evaluate the qualitative results of FTGAN. 
Some examples of the generated images by FTGAN are shown in Figure~\ref{fig:CUB_results}. 
From images shown in Figure~\ref{fig:CUB_results}, we find that our FTGAN are qualified to generate high-quality images with different kinds of bird postures and backgrounds. 
However, it is hard to visually prove the superiority of the proposed FTGAN comparing to the previous work. 
We still need objective criterion to evaluate the quality of the generated images by FTGAN.

Inception Score~\cite{salimans2016improved} is a widely accepted criterion in text-to-image synthesis task. 
To qualitatively examine the images generated by our FTGAN, here we use Inception Score to evaluate the results of FTGAN, as show in Table~\ref{tab:cmp_previous}. Follow the settings in previous work~\cite{reed2016generative,reed2016learning,zhang2017stackgan,zhang2018stackgan++,xu2018attngan}, we generate 10 images for each samples, thus the total number of generated test images is 29330, on which we calculate the Inception Score.

\begin{table}[tb]
	\begin{center}
		\small
		\begin{tabular}{l|l}
			\hline
			Method &Inception Score\\
			\hline
			GAN-INT-CLS~\cite{reed2016generative} &2.88~$\pm$~.04\\
			GAWWN~\cite{reed2016learning}   &3.62~$\pm$~.07\\
			StackGAN~\cite{zhang2017stackgan} &3.70~$\pm$~.04\\
			StackGAN-v2~\cite{zhang2018stackgan++}  &3.82~$\pm$~.06\\
			HDGAN~\cite{zhang2018photographic}  &4.15~$\pm$~.05\\
			AttnGAN~\cite{xu2018attngan} & 4.36~$\pm$~.03\\ 
			MirrorGAN~\cite{qiao2019mirrorgan} & 4.56~$\pm$~.05\\ 
			\hline
			Our FTGAN    &\textbf{4.63~$\pm$~.05}\\
			\hline
		\end{tabular}
	\end{center}
	\vspace{-5pt}
	\caption{Inception Scores by state-of-the-art GAN models~\cite{reed2016generative,reed2016learning,zhang2017stackgan,zhang2018stackgan++,zhang2018photographic,xu2018attngan} and our FTGAN on CUB.}
	\vspace{-5pt}
	\label{tab:cmp_previous} 
\end{table}

From the Table~\ref{tab:cmp_previous} we find that our proposed FTGAN outperforms the state-of-the-art network MirrorGAN, which is also base on AttnGAN. 
MirrorGAN design a global sentence attention to aid the word attention in AttnGAN and utilize the regenerated captions to calculate stream loss replacing the DAMSM loss proposed by AttnGAN, which is far more complex than the image-decoder of FTGAN. 
During the published research, the FTGAN achieves a new state-of-the-art Inception Score in CUB dataset. 
Through the experiments on CUB, we could prove the efficiency of proposed FTGAN.

\subsection{Comparison with previous methods on SCU-Text2face} \label{sec:compare}
\vspace{-5pt}
Since there is no public baseline for text-to-face task, in this section, we will set a baseline for this task by FTGAN.
The test dataset of SCU-Text2face contains 200 face images. 
For each face sample, we generate 10 face images, then evaluate them qualitatively and quantitatively.

\begin{figure*}[tb]
	\begin{center}
		\includegraphics[width=1.0\linewidth]{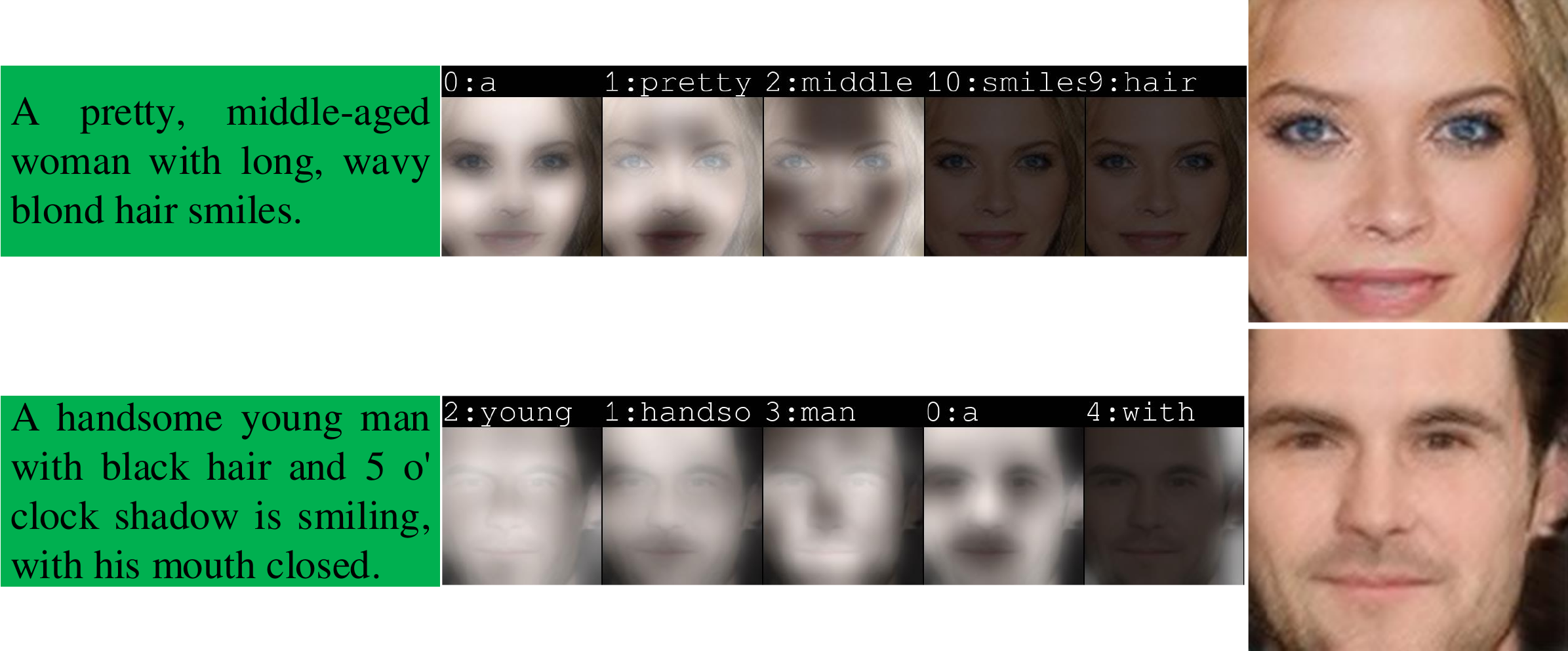}
	\end{center}
	\vspace{-10pt}
	\caption{Two generated examples of text-to-face synthesis. The green boxes in the left are the input sentences.
	}
	\vspace{-10pt}
	\label{fig:face_results}
\end{figure*}

As shown in figure~\ref{fig:face_show}, we could see that FTGAN could generate photo-realistic face images whose quality is close to the ground-truth. 
Besides, the generated images basically match their descriptions. 
For example, all those five face images meet the description of "brown hair" and "slender eyebrows" in the sentence.


Figure~\ref{fig:face_results} shows two examples of the generated face images(right) with its relevant input descriptions(left) and attention maps in generating process(middle). 
For each words in the input sentence, there will be a attention map for it. 
The attention maps serve as input of the inputs in the second and third stage of the image-decoder. 
Showing where the network will focus on for every words when generating images.
The generated attention maps basically match the focusing area of human brain when reacting to those words.
We find that the generated face images are of high consistency with their input sentences. 
For example, the "blond hair" in the first line and "black hair", "mouth closed" in the second line are all presented in their generated face images.

\begin{figure}[tb]
	\begin{center}
		\includegraphics[width=1.0\linewidth]{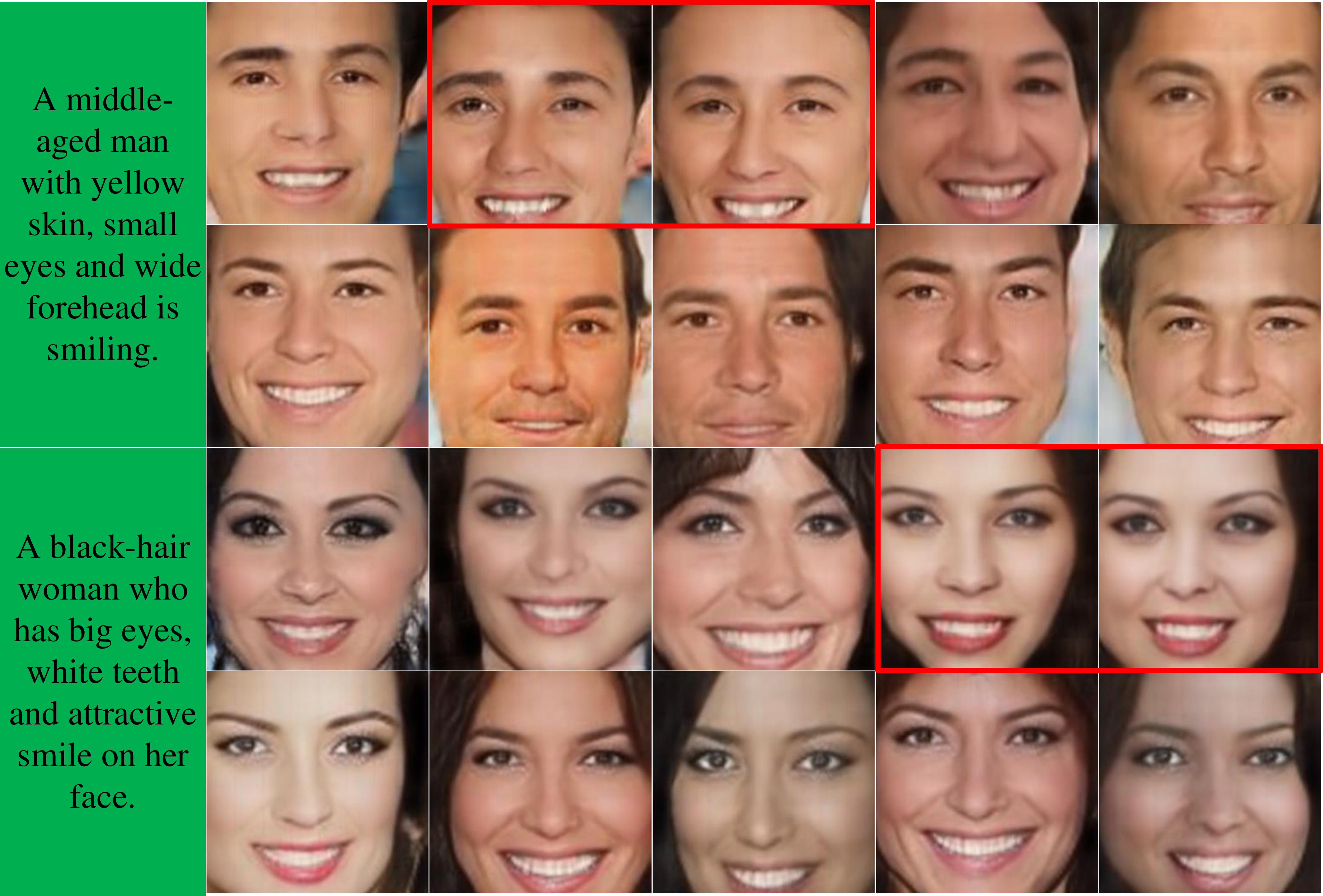}
	\end{center}
	\vspace{-10pt}
	\caption{Comparison between the results of FTGAN and AttnGAN on SCU-Text2face. There are two examples. For each example, the first line is the generated face images by AttnGAN, and the second line is generated by FTGAN.
	}
	\vspace{-10pt}
	\label{fig:face_comparison}
\end{figure}

To prove the superiority of FTGAN, we compare the generated face images of FTGAN and the results of AttnGAN(as shown in figure~\ref{fig:face_comparison}). 
For each input text, the first line is generated by AttnGAN, and the second line is the results of FTGAN. 
As shown in the red boxes, we find that the face images generated by AttnGAN are less diverse than FTGAN. 
And intuitively, our FTGAN is capable to generate higher-quality images than AttnGAN. 

Generally speaking, text-to-image synthesis utilize Inception Score as its criterion. 
To evaluate the networks’ results in CUB, we often use a pre-trained Inception-V3 network which are fine-tuned on CUB to calculate Inception Score. 
However, for face dataset, there are no pre-trained Inception-V3 model. 
So we turned to FID score~\cite{heusel2017gans}, which is another common criterion for evaluating image synthesis, and could be seen as a boosted version of Inception Score.
Besides, in order to evaluate another targets for text-to-face, we refer to two criterion in~\cite{eye2face}. 
Because the final target for text-to-face synthesis is to generate faces similar to their ground-truth just based on the input text, it is a natural idea to judge if the generated face is the same person as ground-truth. 
We utilize FaceNet~\cite{schroff2015facenet} to extract the feature vector of faces, and then calculate the average face semantic distance (FSD) and average face semantic similarity (FSS) between the generated face and ground-truth. 
The formulas of FSD and FSS are shown in formula~\ref{eq:FSD} and formula~\ref{eq:FSS}

\begin{equation}\label{eq:FSD}
FSD =  \dfrac{1}{N}\sum_{i=0}^{N} |Facenet({F_G}_{i})-Facenet({F_GT}_{i})|,
\end{equation}
\begin{equation}\label{eq:FSS}
FSS =  \dfrac{1}{N}\sum_{i=0}^{N} cos(Facenet({F_G}_{i})-Facenet({F_GT}_{i})),
\end{equation}
where $Facenet()$ means using a pre-trained Facenet model to extract a semantic vector of the input face, ${F_G}_{i}$ means one of the generated faces, ${F_GT}_{i}$ means the ground-truth of the synthesized face image. And $cos()$ means calculating the cosine similarity of two vectors.
A higher FSS score and lower FSD score mean the generated face images are more similar to the ground-truth.

\begin{table}[bt]
	\begin{center}
		\small
		\begin{tabular}{l|l|l|l}
			\hline
			Method & FID& FSD & FSS(\%) \\
			\hline
			AttnGAN   &45.56  & 1.269 & 59.28\\
			\hline
			\textbf{FTGAN} &\textbf{44.49}   & \textbf{1.267}   & \textbf{59.41}\\
			\hline
		\end{tabular}
	\end{center}
	\caption{The Fid score and face images' similarity by AttnGAN and our FTGAN on SCU-Text2face.}
	\vspace{-10pt}
	\label{tab:face_comparison} 
\end{table}

The final results are shown in Table~\ref{tab:face_comparison}.
We could find that the FSD value of FTGAN is lower than AttnGAN, consistently, the FSS value of FTGAN is higher than AttnGAN, which means that FTGAN could generate face images which are more similar to the ground-truth than AttnGAN.
Because of the limit of dataset, the face similarity of both networks are not very high, just about 59\%.
However, we find that the generated face images are of high consistency to the input text. 
To our analysis, the main reason is the descriptions of SCU-Text2face are not complex enough.
The descriptions for faces only contain several few attributes (3-5 attributes), which hugely constrains the face similarity between generated face images and its relative ground-truth. 
If there are comprehensive descriptions for every face images, we believe that the face similarity between the synthesized faces and ground-truth will be boosted obviously.

%

\section{Conclusions}
\vspace{-5pt}
This paper proposed a novel text-to-image network FTGAN, which train the text-encoder and image-decoder at the same time. 
Through experiments in the public dataset CUB, FTGAN shows its superiority comparing with the newest state-of-the-art network, achieving 4.63 in Inception Score. 
Though FTGAN have shown its superiority in boosting the quality of generated images comparing to the previous text-to-image synthesis networks, we found this framework are not so stable in the training process. 
In the future, we will try to tackle this problem.

Besides, to fill in the blank in the domain of text-to-face, we build a dataset SCU-Text2face for text-to-face synthesis based on faces in CelebA. 
Every face images in SCU-Text2face have 5 descriptions. 
Based on SCU-Text2face, we set a baseline for text-to-face synthesis task by FTGAN. 
We use FID score to evaluate the image quality of synthesized faces. 
Beside, to evaluate the similarity between generated faces and input text, we calculate the similarity between generated faces and ground-truth to replace it. 
Experiments show that FTGAN could achieve higher-quality images and more similar faces to the ground-truth.
Different from image synthesis on CUB, Oxford102 and COCO, text-to-face generation are more precisely and fixed.
Therefore, to futher improve the quality of generated results, more prior information of face could be added to text-to-face synthesis network.
The task of text-to-face synthesis has huge potentials in public safety domain.
We hope our works could be a good start for this task.

{\footnotesize
\bibliographystyle{ieee}
\bibliography{FTGAN}
}

\end{document}